\documentclass[submission,copyright,creativecommons]{eptcs}
\providecommand{\event}{AMSLO23} 

\usepackage{graphicx, amsmath, amssymb, proof, subcaption, xcolor, oz, braket} 

\usepackage{tikz}
\usepackage{pgfplots}
\usetikzlibrary{trees}
\usetikzlibrary{topaths}
\usetikzlibrary{decorations.pathmorphing}
\usetikzlibrary{decorations.markings}
\usetikzlibrary{matrix,backgrounds,folding}
\usetikzlibrary{chains,scopes,positioning,fit}
\usetikzlibrary{arrows,shadows}
\usetikzlibrary{calc} 
\usetikzlibrary{chains}
\usetikzlibrary{shapes,shapes.geometric,shapes.misc}

\pgfdeclarelayer{edgelayer}
\pgfdeclarelayer{nodelayer}
\pgfsetlayers{background,edgelayer,nodelayer,main}

\tikzstyle{box}=[fill=white, draw=black, shape=rectangle]
\tikzstyle{element}=[fill=white, draw=black, shape=trapezium]
\tikzset{triangle/.style={append after command={
   \pgfextra
        \draw[sharp corners, fill=white, line width = .7pt]%
    ([shift={(-10pt,0)}]\tikzlastnode.south west)-- ([shift={(+10pt,0)}]\tikzlastnode.south east) -- ([shift={(0,+20pt)}]\tikzlastnode.north)-- ([shift={(-10pt,0)}]\tikzlastnode.south west);
   \endpgfextra}}}
\tikzset{spider/.style={append after command={
   \pgfextra
        \draw[fill=white]%
    (\tikzlastnode.center) circle (2mm);
   \endpgfextra}}}
\tikzset{fspider/.style={append after command={
   \pgfextra
        \draw[fill=black]%
    (\tikzlastnode.center) circle (2mm);
   \endpgfextra}}}

\tikzstyle{arrow}=[draw=black, fill=none]
\tikzstyle{bold arrow}=[ line width=3pt]
\tikzstyle{every picture}=[baseline=(current bounding box).east,scale=0.5,node distance=5mm]
\tikzstyle{none}=[inner sep=0pt]
\tikzstyle{every loop}=[]

\tikzstyle{(null)}=[]
\tikzstyle{plain}=[]

\pgfplotsset{compat=1.17}

\newcommand{\semantics}[1]{[\![ #1]\!]}
\newcommand{\csemantics}[1]{\overline{[\![ #1]\!]}}
\newcommand{\vsemantics}[1]{\limg #1 \rimg}
\renewcommand{\id}{\mathrm{id}}
\newcommand{\U}{\mathcal{U}}
\newcommand{\Rel}{\mathbf{Rel}}
\newcommand{\FdVect}{\mathbf{FdVect}}
\newcommand{\C}{\mathcal{C}}
\renewcommand{\L}{\mathbf{L}}
\newcommand{\sllm}{\mathbf{SLLM}}
\newcommand{\bs}{\backslash}

\newcommand{\tobar}{\pfun} 
\renewcommand{\P}{\mathcal{P}}

\title{DisCoCat for Donkey Sentences}
\author{Lachlan McPheat
\qquad
Daphne Wang\\
\institute{University College London}
}
\def\titlerunning{DisCoCat for Donkey Sentences}
\def\authorrunning{L. McPheat \& D. Wang
}

\begin{document}

\maketitle

\begin{abstract}
    We demonstrate how to parse Geach's Donkey sentences in a compositional distributional model of meaning. 
    We build on previous work on the DisCoCat (Distributional Compositional Categorical) framework, including extensions that model discourse, determiners, and relative pronouns.
    We present a type-logical syntax for parsing donkey sentences, for which we define both relational and vector space semantics.
\end{abstract}

\section{Background}
    
    \subsection{Donkey sentences}\label{sec:donkeySentences}
    Montague semantics is a compositional method to translate the semantics of written language into first order logic.
    As a simple example one can understand the meaning of the sentence ``\textit{(all) dogs eat snacks}'' as $\forall x, y.\text{dogs}(x)\wedge \text{snacks}(y)\implies \text{eats}(x,y)$.
    However, when translating the meaning of the sentence ``\textit{Every farmer who owns a donkey beats it}'', the variable representing the donkey cannot be bound by the existential quantifier coming from the determiner `\textit{a}'. 
    This issue was studied by Geach \cite{Geach1962}, using it as a counterexample to the scope of Montague semantics.

    Many have created systems that form semantic representations of donkey sentences, to name a few we have dynamic predicate logic \cite{DPL}, where the binding rules of quantifiers in first order logic are relaxed, discourse representation theory \cite{Kamp2011} where an collection of `discourse referents' keep track of individuals' mentions and are identified to keep track of references, 
    as well as an approach using dependent type theory \cite{luo-2021-donkey}, exploiting dependent sums to differentiate between ambiguous readings of donkey sentences.
    
    However, none of the models mentioned above are type-logical grammars which poses the question whether it is possible to parse donkey sentences and form usable representations of them using type logical grammars?
    We propose to model donkey sentences using (an extension of) Lambek calculus, $\L$.
     
    In the following section, we explain how a type-logical analysis of natural language works, and in sections \ref{sec:relPro},\ref{subsec:generalisedQuantifiers},\ref{sec:DiscourseCat} how to extend it to model more exotic linguistic phenomena, culminating in a parse of a donkey sentence.
    Then we introduce relational semantics and vector space semantics of the extended Lambek calculus in sections \ref{subsec:Rel} and \ref{sec:VSS} respectively, demonstrating how donkey sentence is interpreted as a relation or as a linear map.

    \subsection{Compositional Distributional Models of Meaning}
    
    The framework we use for analysing donkey sentences is compositional distributional models of meaning, also known as DisCoCat (distributional-compositional-categorical).
    Such models combine compositional models of language with distributional models of meaning, such as (neural) language models, although we will not be studying distributional semantics of donkey sentences in this paper. 
    There are several choices of compositional structure such as combinatory categorial grammar (CCG) \cite{CCG}, pregroup grammar \cite{Lambek08} and Lambek calculus, $\L$ \cite{Lambek58}. 
    However these choices of compositional structure turn out to generate the same DisCoCat \cite{Coeckeetal2013, yeung-kartsaklis-2021-ccg}.
    We proceed using Lambek calculus as our syntactic structure, as we find its sequent calculus presentation easier to manipulate, and has a neater categorical semantics.

    Lambek calculus, $\L$ also known as multiplicative intuitionistic linear logic without exchange, is a logic defined over some set of atomic symbols which are chosen to represent grammatical types, for example $n$ for nouns (`\textit{dog}', `\textit{snacks'}), $np$ for noun phrases (`\textit{the dog}', `\textit{John}') and $s$ for sentences (`\textit{Dogs eat snacks}', `\textit{John sleeps}').
    The full set of $\L$-formulas is the free algebra generated over the set of atoms over the three connectives $\bs, /, \bullet$, where the two slashes are implications and $\bullet$ is concatenation (multiplicative conjunction).
    The reason we have two implications is that we require concatenation to be noncommutative ( $A\bullet B\neq B\bullet A$), forcing modus ponens to take two forms, one where the antecedent is on the right of the implication, and one where it is on the left.
    These implications allow us to type functional words, like verbs and adjectives, in a way that preserves word order.
    To illustrate, in English, the concatenation ``\textit{blue car}'' is a noun phrase, but ``\textit{car blue}'' is not, since we require adjectives to be on the left of nouns.
    Hence adjectives are typed $np/n$, meaning that we require an $n$ formula on the right of the adjective to form a noun phrase.
    Similarly intransitive verbs and verb phrases are typed $np\bs s$, and transitive verbs are $np\bs s /np$.

    We define $\L$-sequents as tuples $(\Gamma, A)$, denoted $\Gamma \to A$, where $\Gamma$ is a list of formulas $\{A_1,A_2,\ldots, A_n\}$, and $A$ is a single formula.
    A sequent $\Gamma \to A$ asks whether one can derive $A$ from $\Gamma$ using the rules of $\L$, presented in figure \ref{fig:Lrules}.
    \begin{figure}[htb]
        \centering
            \begin{equation*}
                \begin{array}{cccc}
            \infer[\bs_L]{\Gamma,\Sigma,A\bs B,\Delta \to C}{\Sigma \to A & \Gamma, B, \Delta\to C} 
            &
            \infer[/_L]{\Gamma, B/A,\Sigma,\Delta \to C}{\Sigma\to A & \Gamma,B,\Delta \to C}
            &
            \infer[\bullet_L]{\Gamma, A\bullet B,\Sigma \to C}{\Gamma, A,B,\Sigma\to C}
            &
            \infer[Axiom]{A\to A}{}
            \\ &&& \\
            \infer[\bs_R]{\Gamma \to A\bs B}{A,\Gamma \to B}
            &
            \infer[/_R]{\Gamma \to B/A}{\Gamma,A \to B}
            &
            \infer[\bullet_R]{\Gamma,\Sigma \to A\bullet B}{\Gamma\to A & \Sigma \to B}
            &
            \infer[cut]{\Sigma,\Gamma,\Delta \to B}{\Gamma \to A & \Sigma,A,\Delta\to B}
        \end{array}
            \end{equation*}
        \caption{$\L$ rules\label{fig:Lrules}}
    \end{figure}
    A sequent $\Gamma \to A$ is \textbf{derivable} whenever there is a proof of it, that is a tree with $\Gamma \to A$ as the root, and each branch is a rule of $\L$ and the leaves are all instances of the axiom.

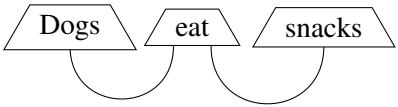
\begin{figure}[htb]
        \centering
        {%
\beginpgfgraphicnamed{DogsEatSnacks}
\begin{tikzpicture}
	\begin{pgfonlayer}{nodelayer}
		\node [style=element] (0) at (-3.25, 2.5) {Dogs};
		\node [style=element] (1) at (0, 2.5) {eat};
		\node [style=element] (2) at (3.5, 2.5) {snacks};
		\node [style=none] (3) at (-0.5, 2) {};
		\node [style=none] (4) at (0.5, 2) {};
		\node [style=none] (5) at (-3.25, 2) {};
		\node [style=none] (6) at (3.5, 2) {};
	\end{pgfonlayer}
	\begin{pgfonlayer}{edgelayer}
		\draw [style=arrow, bend right=90, looseness=1.75] (5.center) to (3.center);
		\draw [style=arrow, bend left=90, looseness=1.75] (6.center) to (4.center);
	\end{pgfonlayer}
\end{tikzpicture}}
\endpgfgraphicnamed}
        \caption{String diagram representing the sentence ``\textit{Dogs eat snacks.}''}
        \label{fig:dogsEatSnacksStr}
    \end{figure}

    For example one can prove that ``\textit{Dogs eat snacks}'' is a sentence by concatenating the $\L$-formulas of the words in the sentence, i.e. $np, np\bs s/np, np$ and asking whether we the sequent $np, np\bs s/np, np \to s$ is derivable in $\L$.
    This is of course the case, as proven in \eqref{eqn:dogsEatSnacksDerivation}.
    \begin{equation}
        \infer[/_L]
        {np, np\bs s/np, np \to s}
        {\infer[]{n\to n}{}
        &
        \infer[\bs_L]{np, np\bs s \to s}
        {\infer[]{np\to np}{}
        &
        \infer[]{s\to s}{}}}
        \label{eqn:dogsEatSnacksDerivation}
    \end{equation}
    By interpreting $\L$ categorically one produces a monoidal biclosed category $\C(\L)$ which comes with a diagrammatic calculus which allows one to draw proofs such as \eqref{eqn:dogsEatSnacksDerivation} as string diagrams, interpreting atoms as trapezia, formulas as strings, and $\bs_L, /_L$-rules as `cups'.
    Without delving into the categorical technicalities, to draw the diagram corresponding to a $\L$-proof, one draws trapezia for each word in the sentence with a string protruding for each $n$ or $np$ occurring in its $\L$-type.
    For example `\textit{dogs}' and `\textit{snacks}' will label trapezia with a single protruding string each, whereas `\textit{eats}' will be a trapezium with two protruding strings.
    One then connects protruding strings by putting a cup on the pair of strings corresponding to the formulas in focus for the $\bs_L$ or $/_L$ rules in the proof.
    This produces a diagram, which is an intuitive representation of proof, akin to the proof nets of linear logic \cite{proofnet}.
    In the case of ``\textit{dogs eats snacks}" we produce the diagram in figure \ref{fig:dogsEatSnacksStr}.
    
    So far we have only discussed the syntactic aspects of DisCoCat, however the main contribution of this area is how it lets us create structured vector representations of meaning.
    This comes from interpreting $\L$ in terms of vector spaces and linear maps, which can be succinctly described categorically as defining a (strongly monoidal closed) functor $\C(\L)\to \FdVect$.
    This means that the formulas $A$ of $\L$ are mapped to finite dimensional vector spaces $V_A$, and $\bs, /, \bullet$ are mapped to the tensor product $\otimes$ and proofs\footnote{technically we map equivalence classes of proofs to linear maps, where we consider proof-theoretic equivalence.}
    of sequents $\Gamma \to A$ are mapped to linear maps $V_\Gamma \to V_A$.
    If you then input distributional vectors into the linear map, it outputs a vector representing the whole compound.

    This framework has been applied to disambiguation and similarity tasks in NLP \cite{KartSadrCoNLL,kartsaklis-sadrzadeh-2013-prior}. 
    However the scope of DisCoCat has been limited by the parsing-capacity of $\L$, which was not able to parse relative pronouns, generalised quantifiers, or \textit{discourses} (written text containing two or more sentences). 
    This has been remedied in extensions of DisCoCat introduced below, starting with how to interpret relative pronouns diagrammatically.

    \subsection{Relative pronouns in DisCoCat}\label{sec:relPro}

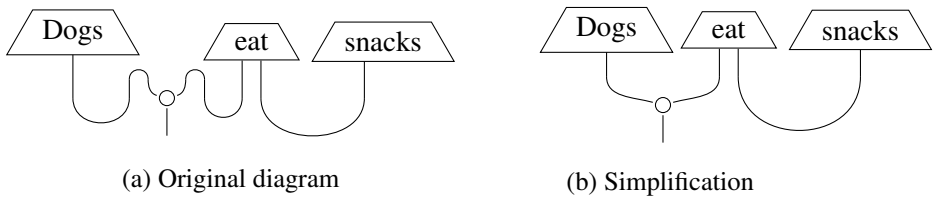
\begin{figure}[htb]
    \centering
        \begin{subfigure}[h]{0.4\textwidth}
         \centering
          {%
\beginpgfgraphicnamed{dogsWhoEatSnacks}
\begin{tikzpicture}
	\begin{pgfonlayer}{nodelayer}
		\node [style=element] (8) at (-3, 1.75) {Dogs};
		\node [style=element] (9) at (1.75, 1.5) {eat};
		\node [style=element] (10) at (5.25, 1.5) {snacks};
		\node [style=none] (14) at (4.75, 0) {};
		\node [style=none] (15) at (2, 0) {};
		\node [style=none] (16) at (-3, 1.25) {};
		\node [style=none] (17) at (1.5, 1) {};
		\node [style=none] (18) at (2, 1) {};
		\node [style=none] (19) at (4.75, 1) {};
		\node [style=none] (20) at (-1.5, .5) {};
		\node [style=none] (21) at (0.5, .5) {};
		\node [style=none] (22) at (-0.5, -1) {};
		\node [style=none] (23) at (-1.5, 0) {};
		\node [style=none] (24) at (0.5, 0) {};
		\node [style=none] (25) at (-1, .5) {};
		\node [style=none] (26) at (0, .5) {};
		\node [style=spider] (27) at (-0.5, 0) {};
		\node [style=none] (28) at (-3, 0) {};
		\node [style=none] (29) at (1.5, 0) {};
	\end{pgfonlayer}
	\begin{pgfonlayer}{edgelayer}
		\draw [style=arrow, bend left=90, looseness=1.25] (14.center) to (15.center);
		\draw [style=arrow] (15.center) to (18.center);
		\draw [style=arrow] (19.center) to (14.center);
		\draw [style=arrow] (24.center) to (21.center);
		\draw [style=arrow] (23.center) to (20.center);
		\draw [style=arrow, bend left=90, looseness=1.75] (20.center) to (25.center);
		\draw [style=arrow, bend left=270, looseness=1.75] (21.center) to (26.center);
		\draw [style=arrow, in=165, out=-90] (25.center) to (27);
		\draw [style=arrow, in=15, out=-90] (26.center) to (27);
		\draw [style=arrow] (27) to (22.center);
		\draw (16.center) to (28.center);
		\draw [bend right=90, looseness=1.50] (28.center) to (23.center);
		\draw [bend left=90, looseness=1.75] (29.center) to (24.center);
		\draw (17.center) to (29.center);
	\end{pgfonlayer}
\end{tikzpicture}}
\endpgfgraphicnamed}
        \caption{Original diagram \label{fig:dogsWhoEatSnacksStr}}
     \end{subfigure}
     \qquad
     \begin{subfigure}[h]{0.2\textwidth}
         \centering
         {%
\beginpgfgraphicnamed{dogsWhoEatSnacks_simplified}
\begin{tikzpicture}
	\begin{pgfonlayer}{nodelayer}
		\node [style=none] (2) at (-1, -1.5) {};
		\node [style=none] (5) at (-2.5, 0.25) {};
		\node [style=none] (6) at (0.5, 0.25) {};
		\node [style=spider] (7) at (-1, -0.5) {};
		\node [style=element] (8) at (-2.5, 1.5) {Dogs};
		\node [style=element] (9) at (0.75, 1.5) {eat};
		\node [style=element] (10) at (4.25, 1.5) {snacks};
		\node [style=none] (14) at (4.25, 0.25) {};
		\node [style=none] (15) at (1, 0.25) {};
		\node [style=none] (16) at (-2.5, 1) {};
		\node [style=none] (17) at (0.5, 1) {};
		\node [style=none] (18) at (1, 1) {};
		\node [style=none] (19) at (4.25, 1) {};
	\end{pgfonlayer}
	\begin{pgfonlayer}{edgelayer}
		\draw [style=arrow, in=165, out=-90] (5.center) to (7);
		\draw [style=arrow, in=15, out=-90] (6.center) to (7);
		\draw [style=arrow] (7) to (2.center);
		\draw [style=arrow, bend left=90, looseness=1.50] (14.center) to (15.center);
		\draw [style=arrow] (15.center) to (18.center);
		\draw [style=arrow] (19.center) to (14.center);
		\draw [style=arrow] (17.center) to (6.center);
		\draw [style=arrow] (16.center) to (5.center);
	\end{pgfonlayer}
\end{tikzpicture}}
\endpgfgraphicnamed}
         \caption{Simplification\label{fig:dogsWhoEatSnacksStr_simplified}}
     \end{subfigure}
     \caption{String diagrammatic parse of ``\textit{Dogs who eat snacks}''}
    \end{figure}

    To analyse the semantics of text containing relative pronouns, the authors of \cite{Sadretal2013Frob,Sadrzadeh2016} introduced ways to represent relative pronouns (e.g. `\textit{who}',`\textit{that}') in the DisCoCat framework.
    This resulted in subject relative pronouns being understood using Frobenius algebras as `internal wiring' for relative pronouns.
    That is, where there would originally be a trapezium labelled `\textit{who}' with three strings protruding, we instead remove the trapezium and replace it with a special diagram of the form \begin{tikzpicture}
	\begin{pgfonlayer}{nodelayer}
		\node [style=none] (0) at (-6, 1.25) {};
		\node [style=none] (1) at (-4, 1.25) {};
		\node [style=none] (3) at (-5, 0.5) {};
		\node [style=none] (5) at (-6, 0.5) {};
		\node [style=none] (6) at (-4, 0.5) {};
		\node [style=none] (7) at (-5.5, 1.25) {};
		\node [style=none] (8) at (-4.5, 1.25) {};
		\node [style=spider] (9) at (-5, 1) {};
	\end{pgfonlayer}
	\begin{pgfonlayer}{edgelayer}
		\draw [style=arrow] (6.center) to (1.center);
		\draw [style=arrow] (5.center) to (0.center);
		\draw [style=arrow, bend left=90, looseness=1.75] (0.center) to (7.center);
		\draw [style=arrow, bend left=270, looseness=1.75] (1.center) to (8.center);
		\draw [style=arrow, in=165, out=-90] (7.center) to (9);
		\draw [style=arrow, in=15, out=-90] (8.center) to (9);
		\draw [style=arrow] (9) to (3.center);
	\end{pgfonlayer}
\end{tikzpicture}
}.
    This lets us parse phrases like ``\textit{Dogs who eat snacks}'' diagrammatically as in figure \ref{fig:dogsWhoEatSnacksStr}.
    The circle is a Frobenius multiplication, which is interpreted as intersection in the relational semantics. 
    This structure will be studied in more detail in section \ref{sec:relSem}.

    \subsection{Generalised quantifiers in DisCoCat} \label{subsec:generalisedQuantifiers}
    In \cite{HedgesSadr2019}, Hedges and Sadrzadeh extended the DisCoCat framework to include the semantics of determiners (e.g. \textit{a}, \textit{every}, \textit{some}, etc.). 
    This framework was based on the theory of \emph{generalised quantifiers} introduced in \cite{generalisedQuantifiers} and gives rise to natural categorical semantics in the category of relations sets and $\Rel$ and in the category of finite dimensional (real) vector spaces and linear maps $\FdVect$. 
    \paragraph{Truth-theoretic semantics}
    We start by defining the standard truth-theoretic models of context free grammars, upon which generalised quantifiers are defined in \cite{generalisedQuantifiers}. 
    From a universe $\U$, which corresponds to the set of all \emph{entities} (i.e. things that can be referred to), each terminal symbol of type $n, np, vp$ is modelled by a subset of $\U$; these can be viewed as unary relations. 
    For example:
    \begin{align*}
        \semantics{\text{dog}} =& \left\{x\in \U ~\middle|~ \text{dog}(x)\right\}\\
        \semantics{\text{sleeps}} =& \left\{x\in \U~\middle|~ \text{sleeps}(x)\right\}
    \end{align*}
    In the case of transitive verbs $v$, we associate instead a binary relation $\semantics{v}\subseteq \U \times \U$, for example:
    \begin{equation*}
            \semantics{\text{eats}} = \left\{(x, y)~\middle|~ x~\text{eats}~y = \text{eats}(x, y)\right\}
        \end{equation*}
    Then, the interpretation of non-terminals are obtained recursively by taking the forward image of the relations defined above. 
    For example, the rules $VP \to V~NP$ and $S\to NP~VP$ are modelled as follows:
    \begin{align*}
            \semantics{V~NP} =& \semantics{v} (\semantics{np})\\
            \semantics{NP~VP} =& \semantics{vp} (\semantics{np}) = \semantics{vp}\cap\semantics{n}
        \end{align*}
    Applied on terminals, this for instance gives:
    \begin{align*}
            \semantics{\text{eats snacks}} =& \left\{x\in \U~\middle|~ \exists s\in \semantics{\text{snacks}}.~ \text{eats}(x, s)\right\}\in \P(\U)\\
            \semantics{\text{dogs sleep}} =& \semantics{\text{dogs}} \cap \semantics{\text{sleep}}\in \P(\U)
    \end{align*}

    We now define the interpretations of determiners following the work of \cite{generalisedQuantifiers}, letting each determiner terminal $d$ be a map $\semantics{d}: \P(\U) \to \P\P(\U)$, for example:
    \begin{align*}
        \semantics{\text{some}}(A) =& \left\{X\subseteq \U ~\middle|~ X\cap A = \emptyset\right\}\\ 
        \semantics{\text{every}}(A) =& \left\{ X \subset \U ~\middle|~ A \subseteq X\right\}
    \end{align*}
    and similarly as with other grammatical rules, we interpret $NP \to Det~N$ as:
    \begin{equation*}
        \semantics{Det~N} = \semantics{d} (\semantics{n})
    \end{equation*}
    This for example gives:
    \begin{equation*}
        \semantics{\text{every dog}} = \left\{X\subseteq \U~\middle|~ \semantics{\text{dogs}}\subseteq X\right\}\in \P\P(\U)
    \end{equation*}

    \paragraph{Relational semantics} In \cite{HedgesSadr2019}, the authors give semantics of a fragment of English containing generalised quantifiers. 
    The most natural way of doing so is to define semantics in the category of sets and relations $\Rel$, where the noun type is taken to be $N = \P(\U)$ and the sentence type to be $S = \{\star\} = I$. 
    As in the standard DisCoCat formalism, each terminal $x$ in $\{n, np\}$ is modelled as a relation%
    \footnote{
    We use the arrow with a vertical bar, $\tobar$, to denote relations, and semantic brackets with a bar, $\csemantics{ \ }$, for relational semantics.
        } 
    $\csemantics{x}: I\tobar N$. 
    In $\Rel$ these are defined as:
    \begin{equation*}
        \csemantics{x} = \left\{(\star, \semantics{x})\right\} \simeq \left\{\semantics{x}\right\}\in \P\P(\U)
    \end{equation*}
    Similarly, we interpret intransitive verbs/verb phrases $vp$, and transitive verbs $v$ as maps $\csemantics{vp}: I\tobar (S\times N)$ and $\csemantics{v}: I\tobar (N\times S\times N)$ respectively as:
    \[
        \csemantics{vp} = \left\{(\star, (\star,\semantics{vp}))\right\} \simeq \left\{\semantics{vp}\right\}\in \P\P(\U)
        \quad \text{  and  }\quad
        \csemantics{v} = \left\{(A, \star, B)~\middle|~ \semantics{v}(B) = A\right\}
    \]
    Finally a determiner $d$ is modelled as a relation $\csemantics{d}: \P(\U) \tobar \P(\U)$, and is defined as:
    \begin{equation*}
        \csemantics{d} = \left\{(A,B)~\middle|~ B\in \semantics{d}(A)\right\}
    \end{equation*}
    This relational structure is endowed with a bialgebra structure $(\scalebox{.5}{\begin{tikzpicture}
	\begin{pgfonlayer}{nodelayer}
		\node [style=spider] (0) at (0, 0) {};
		\node [style=none] (1) at (-0.75, 0.75) {};
		\node [style=none] (2) at (0.75, 0.75) {};
		\node [style=none] (3) at (0, -0.75) {};
	\end{pgfonlayer}
	\begin{pgfonlayer}{edgelayer}
		\draw [bend right=90, looseness=1.75] (1.center) to (2.center);
		\draw (3.center) to (0);
	\end{pgfonlayer}
\end{tikzpicture}}}, \scalebox{.5}{\begin{tikzpicture}
	\begin{pgfonlayer}{nodelayer}
		\node [style=spider] (0) at (0, 0.5) {};
		\node [style=none] (1) at (0, -0.5) {};
	\end{pgfonlayer}
	\begin{pgfonlayer}{edgelayer}
		\draw (1.center) to (0);
	\end{pgfonlayer}
\end{tikzpicture}
}}, \scalebox{.5}{\begin{tikzpicture}
	\begin{pgfonlayer}{nodelayer}
		\node [style=fspider] (0) at (0, 0) {};
		\node [style=none] (1) at (-0.75, -0.75) {};
		\node [style=none] (2) at (0.75, -0.75) {};
		\node [style=none] (3) at (0, 0.75) {};
	\end{pgfonlayer}
	\begin{pgfonlayer}{edgelayer}
		\draw [bend left=90, looseness=1.75] (1.center) to (2.center);
		\draw (3.center) to (0);
	\end{pgfonlayer}
\end{tikzpicture}}}, \scalebox{.5}{\begin{tikzpicture}
	\begin{pgfonlayer}{nodelayer}
		\node [style=fspider] (0) at (0, -0.5) {};
		\node [style=none] (1) at (0, 0.5) {};
	\end{pgfonlayer}
	\begin{pgfonlayer}{edgelayer}
		\draw (1.center) to (0);
	\end{pgfonlayer}
\end{tikzpicture}
}})$ over $N$ where:
    \[
    \begin{array}{crlrl}
        \qquad
        &
        \begin{tikzpicture}
	\begin{pgfonlayer}{nodelayer}
		\node [style=spider] (0) at (0, 0) {};
		\node [style=none] (1) at (-0.75, 0.75) {};
		\node [style=none] (2) at (0.75, 0.75) {};
		\node [style=none] (3) at (0, -0.75) {};
	\end{pgfonlayer}
	\begin{pgfonlayer}{edgelayer}
		\draw [bend right=90, looseness=1.75] (1.center) to (2.center);
		\draw (3.center) to (0);
	\end{pgfonlayer}
\end{tikzpicture}} &=\ \left\{(A, B, A\cap B)~\middle|~ A,B \in \P(\U)\right\}
        &
        \begin{tikzpicture}
	\begin{pgfonlayer}{nodelayer}
		\node [style=spider] (0) at (0, 0.5) {};
		\node [style=none] (1) at (0, -0.5) {};
	\end{pgfonlayer}
	\begin{pgfonlayer}{edgelayer}
		\draw (1.center) to (0);
	\end{pgfonlayer}
\end{tikzpicture}
} &=\ \left\{(\star, \U)\right\}
        \\
        \qquad
        &
        \begin{tikzpicture}
	\begin{pgfonlayer}{nodelayer}
		\node [style=fspider] (0) at (0, 0) {};
		\node [style=none] (1) at (-0.75, -0.75) {};
		\node [style=none] (2) at (0.75, -0.75) {};
		\node [style=none] (3) at (0, 0.75) {};
	\end{pgfonlayer}
	\begin{pgfonlayer}{edgelayer}
		\draw [bend left=90, looseness=1.75] (1.center) to (2.center);
		\draw (3.center) to (0);
	\end{pgfonlayer}
\end{tikzpicture}} &=\ \left\{(A, A, A)~\middle|~ A \in \P(\U)\right\}
        &
        \begin{tikzpicture}
	\begin{pgfonlayer}{nodelayer}
		\node [style=fspider] (0) at (0, -0.5) {};
		\node [style=none] (1) at (0, 0.5) {};
	\end{pgfonlayer}
	\begin{pgfonlayer}{edgelayer}
		\draw (1.center) to (0);
	\end{pgfonlayer}
\end{tikzpicture}
} &=\ \left\{(A, \star)~\middle|~ A\in \P(\U)\right\}
    \end{array}
    \]
    Note that the operations $\left\{\scalebox{.5}{\begin{tikzpicture}
	\begin{pgfonlayer}{nodelayer}
		\node [style=fspider] (0) at (0, 0) {};
		\node [style=none] (1) at (-0.75, -0.75) {};
		\node [style=none] (2) at (0.75, -0.75) {};
		\node [style=none] (3) at (0, 0.75) {};
	\end{pgfonlayer}
	\begin{pgfonlayer}{edgelayer}
		\draw [bend left=90, looseness=1.75] (1.center) to (2.center);
		\draw (3.center) to (0);
	\end{pgfonlayer}
\end{tikzpicture}}}, \scalebox{.5}{\begin{tikzpicture}
	\begin{pgfonlayer}{nodelayer}
		\node [style=fspider] (0) at (0, -0.5) {};
		\node [style=none] (1) at (0, 0.5) {};
	\end{pgfonlayer}
	\begin{pgfonlayer}{edgelayer}
		\draw (1.center) to (0);
	\end{pgfonlayer}
\end{tikzpicture}
}}\right\}$ then act on subsets of $\P(\U)$ while $\left\{\scalebox{.5}{\begin{tikzpicture}
	\begin{pgfonlayer}{nodelayer}
		\node [style=spider] (0) at (0, 0) {};
		\node [style=none] (1) at (-0.75, 0.75) {};
		\node [style=none] (2) at (0.75, 0.75) {};
		\node [style=none] (3) at (0, -0.75) {};
	\end{pgfonlayer}
	\begin{pgfonlayer}{edgelayer}
		\draw [bend right=90, looseness=1.75] (1.center) to (2.center);
		\draw (3.center) to (0);
	\end{pgfonlayer}
\end{tikzpicture}}}, \scalebox{.5}{\begin{tikzpicture}
	\begin{pgfonlayer}{nodelayer}
		\node [style=spider] (0) at (0, 0.5) {};
		\node [style=none] (1) at (0, -0.5) {};
	\end{pgfonlayer}
	\begin{pgfonlayer}{edgelayer}
		\draw (1.center) to (0);
	\end{pgfonlayer}
\end{tikzpicture}
}}\right\}$ act on subsets of $\U$. We also note that $\left\{\scalebox{.5}{\begin{tikzpicture}
	\begin{pgfonlayer}{nodelayer}
		\node [style=fspider] (0) at (0, 0) {};
		\node [style=none] (1) at (-0.75, -0.75) {};
		\node [style=none] (2) at (0.75, -0.75) {};
		\node [style=none] (3) at (0, 0.75) {};
	\end{pgfonlayer}
	\begin{pgfonlayer}{edgelayer}
		\draw [bend left=90, looseness=1.75] (1.center) to (2.center);
		\draw (3.center) to (0);
	\end{pgfonlayer}
\end{tikzpicture}}}, \scalebox{.5}{\begin{tikzpicture}
	\begin{pgfonlayer}{nodelayer}
		\node [style=fspider] (0) at (0, -0.5) {};
		\node [style=none] (1) at (0, 0.5) {};
	\end{pgfonlayer}
	\begin{pgfonlayer}{edgelayer}
		\draw (1.center) to (0);
	\end{pgfonlayer}
\end{tikzpicture}
}}\right\}$ corresponds to the Frobenius structure on $\Rel$, and is therefore defined for all objects in $\Rel$, including $S=I$.
    
    In this semantics, a sentence is then said to be \textbf{true} iff it reduces to the sentence type $S$ (i.e. is grammatically correct) and the corresponding sentence relation $I\tobar I$ is not the empty relation.
    
\begin{figure}[htb]
        \centering
        {%
\beginpgfgraphicnamed{allDogsEatSnacks}
\begin{tikzpicture}
	\begin{pgfonlayer}{nodelayer}
		\node [style=element] (0) at (-3, 1.5) {dog};
		\node [style=element] (1) at (0, 1.5) {eats};
		\node [style=element] (2) at (3.5, 1.5) {snacks};
		\node [style=box] (3) at (-3, 0) {every};
		\node [style=none] (4) at (-3, 1.25) {};
		\node [style=none] (5) at (-3, -0.25) {};
		\node [style=none] (7) at (3.5, 1.25) {};
		\node [style=none] (8) at (-0.25, 1.25) {};
		\node [style=none] (9) at (0.25, 1.25) {};
		\node [style=none] (10) at (-0.25, -0.25) {};
	\end{pgfonlayer}
	\begin{pgfonlayer}{edgelayer}
		\draw (4.center) to (5.center);
		\draw [bend right=90, looseness=1.25] (9.center) to (7.center);
		\draw [bend right=90, looseness=1.25] (5.center) to (10.center);
		\draw (8.center) to (10.center);
	\end{pgfonlayer}
\end{tikzpicture}}
\endpgfgraphicnamed}
        \caption{String diagram representing the sentence ``\textit{every dog eats snacks}'}
        \label{fig:everyDogEatsSnacksStr}
    \end{figure}
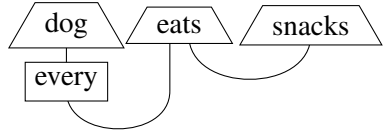

    \paragraph{Example}Consider the sentence ``\textit{Every dog eats snacks}". The diagram associated with the sentence is shown in figure \ref{fig:everyDogEatsSnacksStr}.
    The interpretations of `\textit{Every}', `\textit{dog}', `\textit{eats}' and `\textit{snacks}' are: 
    \begin{align*}
        \csemantics{\text{Every}}=&\{(A,X)\mid A\subseteq X\} : \P(\U) \tobar \P(\U),\\
        \csemantics{\text{dog}}=&\{(*,A)\mid \semantics{\text{Dogs}}=A\} : \{*\} \tobar \P(\U),
        \\
        \csemantics{\text{eats}}=&\{(A,B)\mid A=\semantics{\text{eats}}(B)\}: \{*\} \tobar \P(\U) \times \P(\U),
        \\
        \csemantics{\text{snacks}}=&\{(*,B)\mid \semantics{\text{snacks}}=B\} :\{*\} \tobar \P(\U).
    \end{align*}
    We first obtain the interpretation of $\csemantics{\text{Every dog}}$ by relational composition:
    \begin{equation*}
        \csemantics{\text{Every dog}} = \left\{(\star, X)~\middle|~ \exists A. \star\csemantics{\text{dog}} A \wedge A\csemantics{\text{Every}}X\right\} = \left\{(\star, X)~\middle|~ \semantics{\text{dog}} \subseteq X\right\}: \{\star\}\tobar \P(\U)
    \end{equation*}
    As demonstrated in figure \ref{fig:everyDogEatsSnacksStr} we only need to apply two $\epsilon$-maps to the product of the interpretations above to achieve the semantics of the whole sentence. Note that in figure \ref{fig:everyDogEatsSnacksStr} we also see the diagrammatic treatment of determiners, which are depicted as boxes with both input and output. This is to distinguish them from all other words which are modelled as constants (that is as having no input), which is often denoted by a special shape such as a trapezium or triangle.
    That is:
    \begin{align*}
        &(\epsilon\times\epsilon) \circ (\csemantics{\text{Every dog}} \times \csemantics{\text{eats}} \times \csemantics{\text{snacks}})
        \\
        =& \{* \mid \exists A_1, A_2, B_1, B_2. (A_1,A_2),(B_1,B_2)\in\epsilon \wedge \semantics{dog}\subseteq A_1\wedge A_2=\semantics{\text{eats}}(B_1)\wedge \semantics{\text{snacks}}=B_2\}
        \\
        =& \{* \mid \semantics{\text{dog}}\subseteq \semantics{\text{eats}}(\semantics{\text{snacks}})\}
    \end{align*}
    This sentence is then true whenever the set of entities which eats snacks includes all of the dogs; which is the intended semantics of the sentence ``\textit{Every dog eats snacks}''.

    \paragraph{Free $\FdVect$ construction} As shown in \cite{HedgesSadr2019}, one can construct a model in $\FdVect$ based on the relational semantics. 
    To do so, we take vector space $N$ to have basis  $\{\ket{X}\mid X\in \P(\U)\}$, and the sentence type $S$ to be the monoidal unit $I$, as in $\Rel$, which in $\FdVect$ is the groundfield $I=\mathbb{R}$. 
    Then for each terminal $x\in \{n, np, vp\}$, we define:
    \begin{equation*}
        \vsemantics{x} = \sum_{Y \in \csemantics{x}} \ket{Y}
    \end{equation*}
    And for transitive verbs:
    \begin{equation*}
        \vsemantics{v} = \sum_{X\in \P(\U)} \ket{\semantics{v}(X)}\otimes\ket{X}
    \end{equation*}
    Similarly, for a determiner $d$, we define its interpretation $\vsemantics{d}$ in $\FdVect$ as:
    \begin{equation*}
        \vsemantics{d} = \sum_{Y\in \semantics{d}(X)} \ket{Y}\bra{X}
    \end{equation*}
    And the bialgebra structure is defined accordingly as:
    \[\begin{array}{crlrl}
        \qquad
        &
        \begin{tikzpicture}
	\begin{pgfonlayer}{nodelayer}
		\node [style=spider] (0) at (0, 0) {};
		\node [style=none] (1) at (-0.75, 0.75) {};
		\node [style=none] (2) at (0.75, 0.75) {};
		\node [style=none] (3) at (0, -0.75) {};
	\end{pgfonlayer}
	\begin{pgfonlayer}{edgelayer}
		\draw [bend right=90, looseness=1.75] (1.center) to (2.center);
		\draw (3.center) to (0);
	\end{pgfonlayer}
\end{tikzpicture}} &=\ \sum_{A,B\in \P(\U)} \ket{A\cap B} \left(\bra{A}\otimes \bra{B}\right)
        &
        \begin{tikzpicture}
	\begin{pgfonlayer}{nodelayer}
		\node [style=spider] (0) at (0, 0.5) {};
		\node [style=none] (1) at (0, -0.5) {};
	\end{pgfonlayer}
	\begin{pgfonlayer}{edgelayer}
		\draw (1.center) to (0);
	\end{pgfonlayer}
\end{tikzpicture}
} &=\ \ket{\U}
        \\
        \qquad
        &
        \begin{tikzpicture}
	\begin{pgfonlayer}{nodelayer}
		\node [style=fspider] (0) at (0, 0) {};
		\node [style=none] (1) at (-0.75, -0.75) {};
		\node [style=none] (2) at (0.75, -0.75) {};
		\node [style=none] (3) at (0, 0.75) {};
	\end{pgfonlayer}
	\begin{pgfonlayer}{edgelayer}
		\draw [bend left=90, looseness=1.75] (1.center) to (2.center);
		\draw (3.center) to (0);
	\end{pgfonlayer}
\end{tikzpicture}} &=\ \sum_{A\in \P(\U)} \left(\ket{A}\otimes \ket{A}\right) \bra{A}
        &
        \begin{tikzpicture}
	\begin{pgfonlayer}{nodelayer}
		\node [style=fspider] (0) at (0, -0.5) {};
		\node [style=none] (1) at (0, 0.5) {};
	\end{pgfonlayer}
	\begin{pgfonlayer}{edgelayer}
		\draw (1.center) to (0);
	\end{pgfonlayer}
\end{tikzpicture}
} &=\ \sum_{A\in \P(\U)} \bra{A}
    \end{array}\]
    
    \paragraph{Distributional semantics} We can then define a distributional model on $\FdVect$, which is based of the preceding vector space construction. 
    The difference is that the set of entities $\U$ will be replaced by the set of basis words $\Sigma$. 
    Then, for each words, their vector representation is given by:
    \[\begin{array}{crlrl}
        \qquad
        &
        \vsemantics{n} &=\ \sum_{A\in \P(\Sigma)} c^n_{w} \ket{A}
        &
        \vsemantics{vp} &=\ \sum_{A\in \P(\Sigma)} c^{vp}_A \ket{A}
        \\
        \qquad
        &
        \vsemantics{v} &=\ \sum_{A,B\in \P(\Sigma)} c^v_{A,B} \ket{A}\otimes \ket{B}
        &
        \vsemantics{d} &=\ \sum_{A\in \P(\Sigma)} \sum_{B\in \semantics{d}(A)} c^d_{A,B} \ket{B}\bra{A}
    \end{array}\]
    Note that we are here assuming that the sentence type remains the monoidal unit $I=\mathbb{R}$; there are however other possible choices for sentence spaces, depending on the application \cite{Coeckeetal2010}. 
    It is not always necessary to take into account the whole powerset $\P(\Sigma)$, leading to different ways to interpret the above semantics\cite{HedgesSadr2019}. 
    In \cite{HedgesSadr2019}, the coefficients for terminals in $\{n, np, v, vp\}$ can be obtained from the standard distributional vectors, whilst coefficients $c^d_{A,B}$ in determiners definition is said to quantify the degree to which $d$ of elements of $A$ co-occur with elements of $B$. There are moreover different ways of calculating the different coefficients, regardless of the interpretations. Common choices are probability, conditional probability, likelihood ratio or log likelihood ratio \cite{HedgesSadr2019}. 
    
    \subsection{Discourse in DisCoCat}\label{sec:DiscourseCat}

    A compositional-distributional analysis of discourse has been defined in \cite{mcpheatACT} and improved in \cite{mcpheat2021}, where modal Lambek calculi are employed to model discourse syntactically, and then interpreted in vector spaces to get distributional semantics of discourse.
    The issue with parsing discourse arises in the semantics of \textit{anaphora} and \textit{ellipsis}, or more broadly speaking \textit{reference}.
    Anaphora are words or phrases whose meaning depends on a prior word or phrase.
    A famous class of anaphora are pronouns, such as `\textit{He}' meaning `\textit{John}' in the discourse ``\textit{John sleeps. He snores.}''. 
    Verb phrase-ellipsis is an instance of anaphora occurring when a word refers to a verb-phrase as in ``\textit{Sam plays guitar. Mary does too.}'' where `\textit{does [too]}' refers to the entire VP `\textit{plays guitar}'.
    
    The latest iteration of compositional distributional analysis of discourse uses the Lambek calculus with soft subexponentials, $\sllm$ \cite{Kanovich2020}, which is Lambek calculus with two modalities, $!$ and $\nabla$, allowing $!$-formulas to be copied and $\nabla$-formulas to be permuted. This calculus has a decidable derivation problem once one fixes a global bound on the number of copies in the $!_L$ rule, we call it $k$. The authors also prove a cut-elimination theorem for $\sllm$ \cite{Kanovich2020}.
    
    The copying and permutation are done by adding the rules in figure \ref{fig:sllmRules} to those in figure \ref{fig:Lrules}, in particular the $!_L$ is responsible for copying and the $perm$ rule for permuting. Note that this $!$ modality is not the linear exponential of Girard's linear logic \cite{girard1987}, but rather the soft exponential from Lafont's soft linear logic \cite{Lafont2004}. We also point out the somewhat unorthodox format of the right introduction rules for $!$ and $\nabla$, where the antecedent is restricted to single formulae. The authors of $\sllm$ \cite{Kanovich2020} show that relaxing this to entire structures prohibits cut-elimination.
    \begin{figure}[htb]
        \centering
        \begin{equation*}
            \begin{array}{ccc}
             \infer[!_L]{\Gamma,!A,\Sigma\to B}{\Gamma, \overbrace{A,A, \ldots, A}^{1\leq n \leq k \text{ times }},\Sigma\to B} 
             & 
             \infer[!_R]{!A \to !B}{A \to B}
             &
             \qquad \infer=[perm]{\Gamma, \Sigma, \nabla A, \Delta\to B}{\Gamma,\nabla A,\Sigma,\to B}
             \\ \\
             \infer[\nabla_L]{\Gamma,\nabla A,\Sigma\to B}{\Gamma, A\Sigma\to B}
             & 
             \infer[\nabla_R]{\nabla A \to \nabla B}{A \to B}
             &
        \end{array}
        \end{equation*}
        \caption{$\sllm$-rules}
        \label{fig:sllmRules}
    \end{figure}
    Note that in $\L$ one cannot repeat or move formulas freely as $\L$ formulas represent words, phrases and sentences, which cannot be freely moved or repeated.
    However, according to J\"ager, a type-logical account of reference should consist of copying the referred meaning, and moving one copy to the site of reference\footnote{J\"ager's method manifests syntacticians' view on anaphora, although the debate whether anaphora is a syntactic or semantic structure is an ongoing and lively one within the lingusitics community.} \cite{jager1998multi,jager2006anaphora}.
    For example in the discourse ``\textit{John sleeps. He snores.}'' the meaning of `\textit{John}' is referred to by  `\textit{He}', and so we need to copy the meaning of `\textit{John}' and identify one copy with `\textit{He}'.

    To analyse this in $\sllm$ we decorate referable words with both $!$ and $\nabla$ modalities, so in the case of anaphora `\textit{John}' is typed $!\nabla np$ and in VP-ellipsis `\textit{plays}' is typed $(!\nabla (np\bs s))/np$. 
    We also need to re-type the words doing the referring, rendering pronouns with the new type $\nabla np\bs np$ and VP-ellipsis sites as $\nabla(np\bs s)\bs np\bs s$.
    Here we see that $\nabla$ does not only play the role of movement, but also denotes when formulas are copies, which in turn gives us the $\nabla n\bs n$ typing for pronouns (and similarly for VP-ellipses), namely that such types are `looking' for a copy, and are returning a word of the same type, without the modality.
    This allows for parses of discourses where the references are resolved syntactically.
    Consider for example the derivation of  ``\textit{John sleeps. He snores.}'' in $\sllm$ below:
    
    \begin{equation}
        \infer[!_L]{!\nabla np, np\bs s,\nabla np\bs np, np\bs s \to s\bullet s}
            {\infer[perm]{\nabla np,\nabla np, np\bs s,\nabla np\bs np, np\bs s \to s\bullet s}
            {\infer[\bs_L]{\nabla np, np\bs s,\nabla np,\nabla np\bs np, np\bs s \to s\bullet s}
            {\infer[]{\nabla np \to \nabla np}{}
            &
            \infer[\nabla_L]{\nabla np, np\bs s, np, np\bs s \to s\bullet s}
            {\infer[\bs_L]{np, np\bs s, np, np\bs s \to s\bullet s}
            {\infer[]{np\to np}{}
            &
            \infer[\bs_L]{s, np, np\bs s \to s\bullet s}
            {\infer[]{np\to np}{}
            &
            \infer[\bullet_L]{s, s \to s\bullet s}
            {\infer[]{s\to s}{}
            &
            \infer[]{s\to s}{}}}}}}}}.
            \label{eqn:johnSleepsHeSnores_proof}
    \end{equation}

    There is also a sound string diagrammatic interpretation of this logic, allowing for another intuitive way to represent these derivations for discourses.
    The only addition needed for the string diagrams is that the $!$-modal formulas are now depicted as thick strings, and that $\nabla$-modal formulas have strings that can cross other strings.
    For example, the string diagrammatic parse of ``\textit{John sleeps. He snores.}'' is presented in figure \ref{fig:johnSleepsHeSnores_str}.
    One can even simplify this parse by giving pronouns caps as internal wiring, that is they merely pass along their input\footnote{Technically they also forget that a string can cross other strings, which is not visible in this particular example.}.
    Using this internal wiring we can simplify the diagram in figure \ref{fig:johnSleepsHeSnores_str} to the one in figure \ref{fig:johnSleepsHeSnores_simplified}.

    \begin{figure}[htb]
    \centering
        \begin{subfigure}[h]{0.45\textwidth}
        \centering
        {%
\beginpgfgraphicnamed{johnSleepsHeSnores}
\begin{tikzpicture}
	\begin{pgfonlayer}{nodelayer}
		\node [style=element] (0) at (-8.5, 2.75) {John};
		\node [style=element] (1) at (-4.75, 2.75) {sleeps};
		\node [style=element] (2) at (-1, 2.75) {He};
		\node [style=element] (3) at (2.5, 2.75) {snores};
		\node [style=none] (4) at (-8.5, 2.5) {};
		\node [style=none] (5) at (-4.75, 2.5) {};
		\node [style=none] (7) at (-1.5, 2.5) {};
		\node [style=none] (8) at (-.5, 2.5) {};
		\node [style=none] (9) at (2.5, 2.5) {};
		\node [style=none] (12) at (-4.75, 0.25) {};
		\node [style=none] (14) at (-1.5, 0.25) {};
		\node [style=none] (15) at (-.5, 0.25) {};
		\node [style=none] (16) at (2.5, 0.25) {};
		\node [style=box] (18) at (-8.5, 1.25) {$\pi_2$};
		\node [style=none] (19) at (-9.5, 0) {};
		\node [style=none] (20) at (-7, 0) {};
	\end{pgfonlayer}
	\begin{pgfonlayer}{edgelayer}
		\draw [style=bold arrow] (4.center) to (18);
		\draw [style=arrow, in=75, out=-105, looseness=1.25] (18) to (19.center);
		\draw [style=arrow, in=105, out=-75, looseness=1.50] (18) to (20.center);
		\draw [style=arrow, bend right=90, looseness=1.50] (20.center) to (12.center);
		\draw [style=arrow, bend right=90, looseness=1.25] (19.center) to (14.center);
		\draw [style=arrow] (14.center) to (7.center);
		\draw [style=arrow] (8.center) to (15.center);
		\draw [style=arrow, bend right=90, looseness=1.25] (15.center) to (16.center);
		\draw [style=arrow] (16.center) to (9.center);
		\draw [style=arrow] (12.center) to (5.center);
	\end{pgfonlayer}
\end{tikzpicture}}
\endpgfgraphicnamed}
        \caption{Original version}
        \label{fig:johnSleepsHeSnores_str}
     \end{subfigure}
     \begin{subfigure}[h]{0.45\textwidth}
         \centering
        {%
\beginpgfgraphicnamed{johnSleepsHeSnores_simplified}
\begin{tikzpicture}
	\begin{pgfonlayer}{nodelayer}
		\node [style=element] (0) at (1.5, 2.75) {John};
		\node [style=element] (1) at (5.5, 2.75) {sleeps};
		\node [style=element] (3) at (9.5, 2.75) {snores};
		\node [style=none] (4) at (1.5, 2.5) {};
		\node [style=none] (5) at (5.5, 2.5) {};
		\node [style=none] (8) at (9.5, 2.5) {};
		\node [style=none] (9) at (5.5, 0.25) {};
		\node [style=none] (12) at (9.5, 0) {};
		\node [style=box] (13) at (1.5, 1.25) {$\pi_2$};
		\node [style=none] (14) at (0.5, 0) {};
		\node [style=none] (15) at (2.5, 0) {};
	\end{pgfonlayer}
	\begin{pgfonlayer}{edgelayer}
		\draw [style=bold arrow] (4.center) to (13);
		\draw [style=arrow, in=75, out=-105, looseness=1.25] (13) to (14.center);
		\draw [style=arrow, in=105, out=-75, looseness=1.50] (13) to (15.center);
		\draw [style=arrow, bend right=90, looseness=1.50] (15.center) to (9.center);
		\draw [style=arrow] (12.center) to (8.center);
		\draw [style=arrow] (9.center) to (5.center);
		\draw [bend right=90] (14.center) to (12.center);
	\end{pgfonlayer}
\end{tikzpicture}}
\endpgfgraphicnamed}
        \caption{Simplified version}
        \label{fig:johnSleepsHeSnores_simplified}
     \end{subfigure}
     \caption{String diagrammatic interpretation of ``\textit{John sleeps. He snores.}''}
    \end{figure}

    We note that the analysis provided by $\sllm$ is not complete, and there are possible refinements to make. One is that strictly speaking, anaphora are words referring to \textit{previous} ones, but with the typing provided one can parse ``\textit{*He sleeps. John snores.}". This can easily be fixed by forcing the permutation rule to only allow rightwards permutation, although this would prohibit parsing \textit{cataphora} which are words/expressions referring forwards, which requires leftward permutation. There is a clear trade-off: $\sllm$ which overgenerates but parses both anaphora and cataphora or $\sllm$ with only rightward/leftward permutation which is only able to parse anaphora \textit{or} cataphora. Furthermore the $\nabla_L$ rule lets us forget that a formula was copied, which indeed is too lax for some discourses, for instance it allows the same noun phrase to be both the subject and object of any sentence missing an object. This issue constitutes ongoing work, where the first step to addressing this issue lies in the syntax.
    
\section{Syntax}

    Although no explicit treatment of syntax of generalised quantifiers in terms of Lambek calculus was given, we can simply translate the grammar of \cite{HedgesSadr2019} into a Lambek calculus as follows.
    The only production rule producing determiners is $NP \to Det \ n$ meaning that determiners must be of Lambek type $np/n$ (take a noun on the right and return a noun-phrase). 
    However, as we are interested in referring to quantified noun phrases, we must decorate this $\L$-type with $\sllm$-modalities to ensure the appropriate copying behaviour.
    As noted in section \ref{sec:donkeySentences}, we have a reference occurring when the `\textit{it}' refers to the phrase `\textit{a donkey}'.
    This means that the noun-phrase produced by composing `\textit{a}' and `\textit{donkey}' is referable, forcing the determiner to be typed $(!\nabla np)/n$.

    It is surprisingly useful to cover the whole determiner-type with the $!\nabla$ combination, i.e. typing determiners as $!\nabla(!\nabla (np)/n)$, due to the fact that the projection operations introduced in the semantics below are natural transformations. This means that ``\textit{all (dogs and cats)}" and ``\textit{all dogs and all cats}" would have the same semantics.
    The projectors being natural means that referring to a quantified phrase or referring to a phrase and then quantifying yields the same meaning.

    With this typing we can parse the donkey sentence in $\sllm$, using the usual Lambek typing with the addition of $!\nabla$ on the determiners and $\nabla np\bs np $, as motivated in \cite{mcpheat2021}.
    This gives us the following dictionary:
    \[\small\{\text{Every, a}:\ !\nabla(!\nabla np)\bs n, 
    \text{farmer, donkey}:n,
    \text{owns, beats}: np\bs s/np,
    \text{who}: np\bs np/(np\bs s), 
    \text{it}: \nabla np\bs np
    \}\]
    allowing us to derive the sequent corresponding to ``\textit{Every farmer who owns a donkey beats it}'', that is:
    \[!\nabla(!\nabla np)/ n, n, np\bs np/(np\bs s), np\bs s/np,!\nabla(!\nabla (np)/n), n, np\bs s/np, \nabla np\bs np \to s \]
    For legibility reasons we make our proof shorter by removing the modalities from `\textit{Every}' as we are not referring to `\textit{Every farmer}' in the donkey sentence, giving us the slightly shorter proof in figure \ref{fig:donkeyDerivation}:

    \begin{figure}[htb]
        \centering
        \[\scalebox{.8}{\infer[/_L]{np/ n, n, np\bs np/(np\bs s), np\bs s/np,!\nabla(!\nabla (np)/n), n, np\bs s/np, \nabla np\bs np \to s }
            {\infer[]{n \to n}{}
            &
            \infer[!_L]{np, np\bs np/(np\bs s), np\bs s/np,!\nabla(!\nabla (np)/n), n, np\bs s/np, \nabla np\bs np \to s}
            {\infer[\nabla_L]{np, np\bs np/(np\bs s), np\bs s/np,\nabla(!\nabla (np)/n), n, np\bs s/np, \nabla np\bs np \to s}
            {\infer[/_L]{np, np\bs np/(np\bs s), np\bs s/np,!\nabla (np)/n, n, np\bs s/np, \nabla np\bs np \to s}
            {\infer[]{n\to n}{}
            &
            \infer[!_L]{np, np\bs np/(np\bs s), np\bs s/np,!\nabla np, np\bs s/np, \nabla np\bs np \to s}
            {\infer[perm]{np, np\bs np/(np\bs s), np\bs s/np,\nabla np,\nabla np, np\bs s/np, \nabla np\bs np \to s}
            {\infer[\bs_L]{np, np\bs np/(np\bs s), np\bs s/np,\nabla np, np\bs s/np,\nabla np \nabla np\bs np \to s}
            {\infer[]{\nabla np\to \nabla np}{}
            &
            \infer[/_L]{np, np\bs np/(np\bs s), np\bs s/np,\nabla np, np\bs s/np, np \to s}
            {\infer[]{np\to np}{}
            &
            \infer[\nabla_L]{np, np\bs np/(np\bs s), np\bs s/np,\nabla np, np\bs s \to s}
            {\infer[/_L]{np, np\bs np/(np\bs s), np\bs s/np, np, np\bs s \to s}
            {\infer[]{np \to np}{}
            &
            \infer[/_L]{np, np\bs np/(np\bs s), np\bs s, np\bs s \to s}
            {\infer[]{np\bs s\to np\bs s}{}
            &
            \infer[\bs_L]{np, np\bs np, np\bs s \to s}
            {\infer[]{np\to np}{}
            &
            \infer[\bs_L]{np, np\bs s \to s}
            {\infer[]{np \to np}{}
            &
            \infer[]{s \to s}{}}}}}}}}}}}}}}}
        \]
        \caption{Donkey sentence $\sllm$-derivation}
        \label{fig:donkeyDerivation}
    \end{figure}
    
\section{Relational semantics of a donkey sentences using DisCoCat}\label{sec:relSem}

    \subsection{Fock spaces in sets and relations}\label{subsec:Rel}

    From the work of \cite{mcpheat2021}, introduced in section \ref{sec:DiscourseCat}, we see that for us to have a semantics of anaphora and ellipsis, we require an endofunctor $M$ with projectors $(\pi_n: M\to \id^n)_{1\leq n \leq k}$ on our semantic category. 
    As we are interested in semantics in $\Rel$, we need to find such structure in $\Rel$, which thankfully is quite easy given that $\Rel$ has finite biproducts, namely disjoint unions.
    Hence we can define the Fock space on $\Rel$ as the endofunctor $F$ mapping
    \[\begin{array}{cc}
        \text{objects } : X \mapsto \bigsqcup_{i=1}^k X^i 
        &
        \text{ and morphisms } : (R:X\tobar Y) \mapsto (\bigsqcup_{i=1}^k \sigma_i(R^i) : \bigsqcup_{i=1}^k X^i \tobar \bigsqcup_{i=1}^k Y^i)
    \end{array}\]
    where $\sigma_i(R^i)$ is the image of $R^i$ under the ($\mathbf{Set}$-)isomorphism \((X\times Y)^i \cong X^i \times Y^i \),
    which is necessary to ensure that the new relation $F(R)$ is well-typed, i.e. that $F(R) \subseteq F(X)\times F(Y)$.
    To get an idea of how $F(R)$ works, consider a tuple\footnote{
    Note that elements in $F(X)$ are tuples of length $1,\ldots,k$ which we denote as pairs $(\vec{x},n)$, meaning $\vec{x}=(x_1,x_2, \ldots, x_n)$.
    } 
    $(\vec{x},n,\vec{y},m)\in F(X)\times F(Y)$. We have that \[(\vec{x},n,\vec{y},m)\in F(R) \iff n=m \text{ and } (x_i,y_i)\in R \text{ for } i=1,\ldots, n.\]

    


    
    We define the projection maps $\pi_n:F\to \id^n$ at a set $X$ to be relations 
    \[\pi_{n,X} : F(X) \tobar X^n := \{((x_1,x_2,\ldots,x_m,m,x'_1,\ldots, x'_n) \mid n=m \text{ and } x_i=x'_i 
    \text{ for } i=1,\ldots, n\}.\]
   For example, in the case of pronominal anaphora we often make use of a $\pi_{2,X}$ projection. In $\Rel$ such a relation will contain tuples of the form $(((x_1,x_2),2),x_1,x_2)$. 
   Note that $\pi_n$ are natural transformations, so one can quantify-then-project or project-then-quantify without getting different derivations.
    \subsection{Parsing of a donkey sentence}\label{subsec:parse}
    
    \begin{figure}[htb]
        \centering
        {%
\beginpgfgraphicnamed{layeredParse}
\begin{tikzpicture}style={
  trapezium, trapezium angle=67.5, draw,
  inner sep=0pt, outer sep=0pt,
  minimum height=1.81mm, minimum width=0pt
}
	\begin{pgfonlayer}{nodelayer}
		\node [style=element] (0) at (-6.75, 4.5) {every farmer};
		\node [style=element] (1) at (-2, 4.5) {owns};
		\node [style=element] (2) at (2, 4.5) {a donkey};
		\node [style=element] (3) at (6.5, 4.5) {beats};
		\node [style=none] (4) at (-6.75, 4) {};
		\node [style=none] (5) at (2, 4) {};
		\node [style=none] (6) at (-2.5, 4) {};
		\node [style=none] (8) at (-1.5, 4) {};
		\node [style=none] (9) at (6, 4) {};
		\node [style=none] (11) at (7, 4) {};
		\node [style=spider] (12) at (-4.75, 2) {};
		\node [style=none] (13) at (-4.75, -2) {};
		\node [style=box] (14) at (2, 2) {$\pi_2$};
		\node [style=none] (15) at (0, 0) {};
		\node [style=none] (17) at (4, 0.25) {};
		\node [style=none] (18) at (6, 0.25) {};
		\node [style=none] (19) at (4, -0.75) {};
		\node [style=none] (20) at (6, -0.75) {};
		\node [style=none] (21) at (-1, -2) {};
		\node [style=none] (23) at (-1.5, 0) {};
		\node [style=none] (24) at (6, -2) {};
		\node [style=none] (25) at (7, -2) {};
		\node [style=none] (26) at (-10, 3) {};
		\node [style=none] (27) at (9, 3) {};
		\node [style=none] (28) at (-10, 0.75) {};
		\node [style=none] (29) at (9, 0.75) {};
		\node [style=none] (30) at (-10, -1.5) {};
		\node [style=none] (31) at (9, -1.5) {};
	\end{pgfonlayer}
	\begin{pgfonlayer}{edgelayer}
		\draw [style=arrow, in=-180, out=-90] (4.center) to (12);
		\draw [style=arrow, in=-90, out=0] (12) to (6.center);
		\draw [style=arrow] (12) to (13.center);
		\draw [style=bold arrow] (5.center) to (14);
		\draw [style=arrow, in=90, out=-120] (14) to (15.center);
		\draw [style=arrow, in=90, out=-90, looseness=0.75] (17.center) to (20.center);
		\draw [style=arrow, in=-90, out=90, looseness=0.75] (19.center) to (18.center);
		\draw [style=arrow] (18.center) to (9.center);
		\draw [style=arrow, bend right=90, looseness=1.50] (13.center) to (21.center);
		\draw [style=arrow, in=-90, out=90, looseness=0.50] (21.center) to (19.center);
		\draw [style=arrow, bend left=90, looseness=1.50] (15.center) to (23.center);
		\draw [style=arrow] (23.center) to (8.center);
		\draw [style=arrow] (20.center) to (24.center);
		\draw [style=arrow, bend right=90, looseness=2.00] (24.center) to (25.center);
		\draw [style=arrow] (25.center) to (11.center);
		\draw [style=dashed] (26.center) to (27.center);
		\draw [style=dashed] (28.center) to (29.center);
		\draw [style=dashed] (30.center) to (31.center);
		\draw [style=arrow, in=90, out=-60] (14) to (17.center);
	\end{pgfonlayer}
\end{tikzpicture}}
\endpgfgraphicnamed}
        \caption{First stage of parsing the donkey sentence}
        \label{fig:donkeySentence1}
    \end{figure}
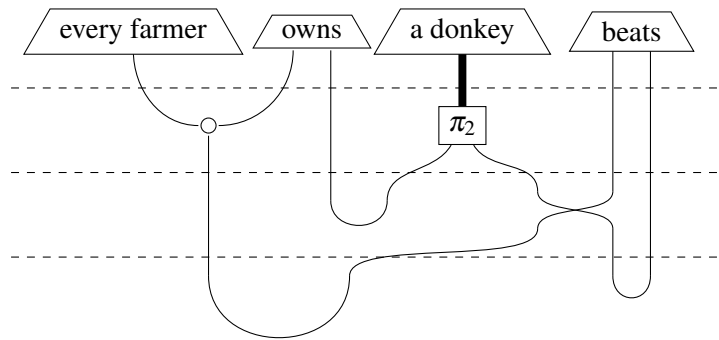

    The diagram in figure \ref{fig:donkeySentence1} represents a relation \(\{*\} \tobar \{*\}\) (i.e. a sentence) with dashed lines splitting it into parts allowing us to describe the parse step-by-step. 
    We have also pre-composed the determiners with the nouns to shorten the parse, and refer to \cite{HedgesSadr2019} for details on this.
    To understand what this relation is in detail, we parse the string diagram as follows.
    The top level, i.e. the input of the sentence, is a relation 
    $\{*\} \tobar \P(\U)^3 \times F(\P(\U)) \times \P(\U)^2$    
    given by 
    $\csemantics{\text{every farmer}} \times \csemantics{\text{owns}} \times F(\csemantics{\text{a donkey}})\times \csemantics{\text{beats}}$,
    which is simply the set 
    $
       \{(*, F_1, F_2, D_1, \vec{D}, F_3, D_3) \mid \phi_1(F_i,D_j,\vec{D})
       \}
    $
    where $\phi_1$ is the predicate 
    \[ 
        \semantics {\text{farmer}} \subseteq F_1 \wedge (F_2, D_1) \in \semantics{\text{owns}} \wedge B\cap\semantics{\text{donkey}}\neq \emptyset \wedge (F_3,D_3)\in\semantics{\text{beats}}
    \]
    We use the notation that domain variables for sets pertaining to farmers are $F_i$ and $D_j$ for donkeys, and $\vec{D}$ for elements of the in the Fock set $F(\csemantics{\text{a donkey}})$ and corresponding codomain variables are denoted by $F'_i,D'_j$.

    The second level of figure \ref{fig:donkeySentence1} is a relation $\P(\U)^3\times F(\P(\U))\times \P(\U)^2 \tobar \P(\U)^6$ which we get by applying $\mu$ and $\pi_{2,\P(\U)}$ maps, giving us the relation $\mu \times \id \times \pi_{2,\P(\U)} \times \id \times \id$. 
    This relation is given by the set 
        \begin{align*}
        \{(F_1,F_2,D_1,\vec{D}, F_3,D_2, F'_1, D'_1,D'_2,F'_2,D'_3,D'_4)
        \mid& \phi_2(F_i,D_j,\vec{D}, F'_i, D'_j)
        \}
        \\ \cong
        \{(F_1,F_2,F_3,D_1,\vec{D},D_2, F',D'_1,D'_2)
        \mid&
        F_1 \cap F_2 = F' 
        \text{ and }
        D=D'_1=D'_2\},
        \end{align*}
    where $\phi_2$ is the predicate $(F_1,F_2,F'_1)\in \mu \wedge (D_1,D'_1), (F_3,F'_2),(D_2,B'4) \in \id \wedge (\vec{D},D'_2,D'_3) \in \pi_{2,\P(\U)}$.
    
    Composing the first and second levels gives us a relation $\{*\} \tobar \P(\U)^6$ given by the set:
    \begin{align*}
        \{(F'_1,D'_1,D'_2,D'_3,F'_2,D'_4)
        \mid & 
        \exists F_1, F_2, D_1, \vec{D}, F_3,D_3.
        \phi_x((F_i,D_j,\vec{D}, F'_i, D'_j)) ,\text{ for } x=1,2
        \}
    \end{align*}

    The third level of figure \ref{fig:donkeySentence1} only contains a cup ($\epsilon$) and a swap ($\sigma$), giving us the relation $\id\times \epsilon \times \sigma\times \id$ which is a relation $\P(\U)^6 \tobar \P(\U)^4$, given by the following set:
    \begin{align*}
        \{(F_1,D_1,D_2,D_3,F_2,D_4, F'_1, F'_2, B'1, D'_2)
        \mid & \phi_3(F_i,D_j,\vec{D}, F'_i, D'_j)
        \}
        \\\cong 
        \{(F_1,D_1,D_2,F_2,D_3, F'_1, D'_1)
        \mid &
        D_1 = D_2, 
        (D_3,F_2,F'_1,D'_1)\in \sigma,
        \},
    \end{align*}
    where $\phi_3$ predicates $(F_1,F'_1), (D_4,D'_2)\in \id \wedge  (D_1,D_2)\in \epsilon \wedge (D_3,F_2,F'_2,D'_1)\in \sigma$.
    
    When composed with the two layers above we get the following relation $\{*\}\tobar\P(\U)^4$:
    \begin{align*}
        \{
            (F'_1, F'_2, D'_1, D'_2) 
            \mid &
            \exists F_1, D_1, D_2, D_3,F_2, D_4. \phi_x(F_i,D_j,\vec{D}, F'_i, D'_j), \text{ for } x =1,2,3 
        \}
    \end{align*}
    
    The final section of figure \ref{fig:donkeySentence1} contains only two cups ($\epsilon$) giving us the relation $\epsilon \times \epsilon : \P(\U)^4 \tobar \{*\}$, consisting of the set 
    \[\{(F_1,F_2,D_1,D_2)\mid (F_1,F_2)\in \id, (D_1,D_2)\in \id\} \cong \{(A,B)\},\]
    which when composed with the previous relation gives us the entire sentence relation $\alpha: \{*\} \tobar \{*\}$ which when applied to the input:
    \[
        \alpha \left(\csemantics{every}\circ \csemantics{farmer} \times \csemantics{owns} \times \csemantics{a}\circ \csemantics{donkey}\times \csemantics{beats} \csemantics{it} \right) 
    \]
    gives us the set:
    \begin{align*}
        \{*,*\mid & \exists F_1,F_2,F_3, D_1,D_2.\semantics{farmer}\subseteq F_1,
        F_1\cap F_2 = F_3,
        \semantics{donkey}\cap D_1 \neq \emptyset,\\
        &\semantics{donkey}\cap D_2 \neq \emptyset,
        F_2 =\semantics{owns}(D_1),
        F_3 = \semantics{beats}(D_2)
        \}
    \end{align*}

\subsection{Vector semantics of donkey sentences using DisCoCat}\label{sec:VSS}

    The Fock space construction described above leads to a free vector space interpretation using the construction of \cite{HedgesSadr2019}. 
    The semantics of sentences without reference are the same as the ones defined in Section \ref{subsec:generalisedQuantifiers}. 
    We recall the the endofunctor $F:\FdVect\to \FdVect$ of \cite{mcpheat2021} below and note its similarity to the one in section \ref{subsec:Rel}.
    \[\begin{array}{cc}
        \text{objects } : V \mapsto \bigoplus_{i=1}^k V^{\otimes i}
        &
        \text{ morphisms } : (m:V\to W) \mapsto (\bigoplus_{i=1}^k m^{\otimes i} : \bigoplus_{i=1}^k V^{\otimes i} \to \bigoplus_{i=1}^k W^{\otimes i})
    \end{array}\]
    Similarly, we define the projection operators $\pi_{n,V}: FV \to V^{\otimes n}$ as:
    \[
        \pi_{n, V}\left(\bigoplus_{i=1}^k \overrightarrow{x_i}\right) = \overrightarrow{x_n}
    \]
    where $\overrightarrow{x_i}\in V^{\otimes i}$ for each $i = 1, \ldots, k$.
    
    By applying the definitions of morphisms from Section \ref{subsec:generalisedQuantifiers}, as well as the procedure described in Section \ref{subsec:parse}, it can be shown that the parse of the donkey sentence ``\textit{Every farmer who owns a donkey beats it}'' in vector semantics is given by:
    \begin{equation*}
        \sum_{F_1|\semantics{farmer}\subseteq F_1} \sum_{D_1|D_1\cap\semantics{donkey}\neq \emptyset} \sum_{D_2|D_2\cap\semantics{donkey}\neq \emptyset} \braket{F_1\cap\semantics{owns}(D_1)|\semantics{beats}(D_2)}
    \end{equation*}
    This is indeed true (i.e. does not evaluate to $0\in \mathbb{R}$), iff the parse in $\Rel$ also evaluates to true.


\section{Conclusion and Outlook}

\paragraph{Summary of results} We have combined three extensions to compositional distributional models of meaning allowing for modelling relative pronouns, generalised quantifiers and anaphora, yielding a model that can parse Geach's donkey sentences. We have provided a syntax in form of $\sllm$ to allow for a type-logical parsing of donkey sentences, as well as a relational semantics for $\sllm$, in particular a relational version of Fock spaces. The semantics of the donkey sentence has been derived both as a relation and and as a linear map.

\paragraph{Future work}Using this framework we are able to implement a distributional representation of quantified sentences with reference from empirical data mined in corpora (e.g BNC, ukWaC, \ldots). This implementation could be use to measure the ``degree of truth'' of sentences with referents and quantifiers, and could subsequently be used for NLP tasks. In addition, it would be interesting to model other ellipses and anaphoric phenomena such as sluicing or interaction with conjunction and disjunction. 

Furthermore, it remains to derive the alternate `strong' reading of the donkey sentence using this distributional compositional framework. That is, the reading where one interprets `a donkey' universally. This is studied in \cite{DPL,luo-2021-donkey}, where it is concluded that this is a semantic condition. The compositional-distributional framework we use does not encode the universal interpretation of `a' in this context, since we were trying to combine three existing models into one that is able to parse donkey sentences and so have not tried to add to any of the models as such. However this could perhaps be solved by allowing ambiguous determiners to be have disjuntion of semantics, allowing for multple meanings arising from the same parse. This constitutes further work.
	
\bibliographystyle{eptcs}
\bibliography{references.bib}    
\end{document}